# A 2D Georeferenced Map Aided Visual-Inertial System for Precise UAV Localization

MAO Jun, ZHANG Lilian, HE Xiaofeng, QU Hao, HU Xiaoping

*Abstract* — Precise geolocalization is crucial for unmanned aerial vehicles (UAVs). However, most current deployed UAVs rely on the global navigation satellite systems (GNSS) for geolocalization. In this paper, we propose to use a lightweight visual-inertial system with a 2D georeferenced map to obtain accurate geodetic positions for UAVs. The proposed system firstly integrates a micro inertial measurement unit (MIMU) and a monocular camera to build a visual-inertial odometry (VIO) to consecutively estimate the UAV's motion states and reconstruct the 3D position of the observed visual features in the local world frame. To obtain the geolocation, the visual features tracked by the odometry are further registered to the 2D georeferenced map. While most conventional methods perform image-level aerial image registration, we propose to align the reconstructed 3D points with the map, and then use the registered 3D points to relocalize the vehicle in the geodetic frame, which helps to improve the geolocalization accuracy. Finally, a pose graph is deployed to fuse the geolocation from the point registration and the local navigation result from the visual-inertial odometry (VIO) to obtain smooth and drift-free geolocalization results. We have validated the proposed method by installing the sensors to a UAV body rigidly and have conducted two real-world flights in different environments with unknown initials. The results show that the proposed method can achieve less than 4m position error in flight at about 100m high and less than 9m position error in flight at about 300m high.

## I. INTRODUCTION

The navigation system is the key component that supports the UAVs to perform various tasks such as search and rescue, surveying and mapping. Currently, most UAVs integrate an INS with the GNSS to obtain precise motion state estimation. However, the GNSS signals are subject to interference and multipath propagation, which can cause the navigation precision to fall dramatically; such degradation is more notable for the systems that integrate with low precision MIMUs.

With the development of computer vision, visual navigation techniques have been widely deployed in UAVs. Well-developed visual navigation algorithms have achieved notable precision. The field of visual navigation comprises two main approaches: relative visual localization (RVL) and absolute visual localization (AVL) [1]. The RVL algorithms include visual odometry (VO) and visual simultaneous localization and mapping (VSLAM). VO estimates the relative motion between two consecutive frames or keyframes,

*Research supported by the National Nature Science Foundation of China under Grant 62103430, 62103427 and 62073331.

All authors are with the College of Intelligence Science and Technology, National University of Defense Technology, Changsha, China. Corresponding author to maojun12@nudt.edu.cn.

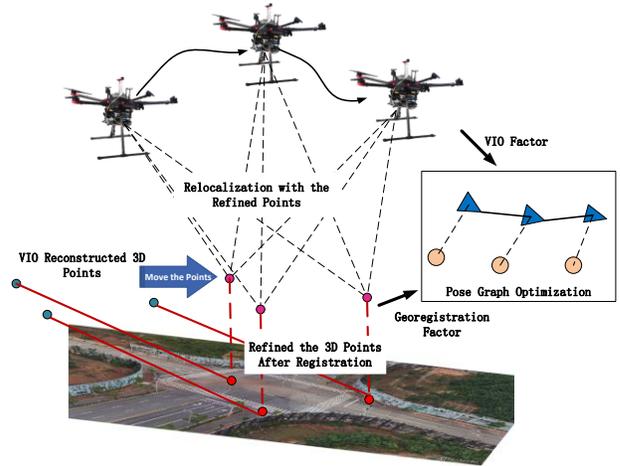

Figure 1. The proposed method uses VIO reconstructed 3D points to perform georegistration, and uses a pose graph to fuse the VIO and geolocalization results to achieve smooth, drift-free and precise localization performance.

and then integrates the relative motion to obtain the position and attitude of the agent [2]. However, the errors in each relative motion estimation accumulate with the integration process. To reduce the drifts, IMUs have been combined with cameras to form visual-inertial odometry [3]. VO or VIO also acts as a component in VSLAM systems, which jointly estimates the vehicle's position and reconstructs the 3D position of the observed landmarks [4][5]. While the vehicle returns to an explored area, VSLAM can perform loop closure to reduce the accumulated drift from odometry and adjust the constructed map.

The development of RVL techniques has supported various UAV applications. However, visual odometry suffers from accumulated drifts; although loop closure in VSLAM can reduce the drifts, it requires the vehicle to travel back to previously explored areas, which is not efficient and safe in some situations. Furthermore, in many applications, the UAVs are desired to navigate to desired locations represented by latitudes and longitudes; however, without known initial position and orientation, RVL is not capable to locate the vehicle in the geodetic frames.

In contrast to RVL techniques, AVL compares the camera images with a map to localize the vehicle. When the map is aligned with the geodetic frame, the output of AVL is the geodetic location; moreover, there is no integration operation in AVL, which brings in the advantage of drift-free position estimation. The availability of georeferenced maps from online platforms, such as Google Earth and Bing map, has also motivated the research of AVL [6][7][8].

In UAV applications, template matching and feature matching are two commonly used approaches to register the aerial images with the georeferenced map image to achieve AVL performance [1]. However, most existing work only performs image-level registration, where the camera images are treated as a patch from the map image. To obtain downward-looking views similar to the map, gimbals have been used to keep the onboard camera facing to the ground [6][7][9][10] and known initial positions [6][7][8][9][11][12] are required to achieve good performances. However, current UAVs are usually equipped with multiple sensors distributed on different parts of the UAV body (such as the front view camera for obstacle avoidance and the downward-looking camera for localization), using a gimbal installation makes it hard to integrate the sensors at different parts. Moreover, known initial positions are not easy to obtain; in many challenging situations, the vehicle has to localize itself with unknown initials or kilometer-scale initial uncertainties.

In this paper, we provide a novel approach that combines an RVL and an AVL technique to obtain an accurate and smooth estimation of the UAV's geolocations. The proposed algorithm consists of three components: a visual-inertial odometry, a feature-based point registration component and a pose graph fusion component. Unlike most conventional geolocalization methods that focus on image-level registration [6][7][8][9][12] or utilize a 3D georeferenced model [11], the proposed method uses a 2D map and makes use of the VIO reconstructed 3D points to perform georegistration. Moreover, we propose to use a pose graph to fuse the VIO and the georegistration results to obtain smooth and drift-free localization in the geodetic frame. The proposed approach is illustrated in Fig.1.

The main contributions of this work are as follows:
1. We propose a novel georegistration method that aligns the VIO reconstructed 3D points to the 2D georeferenced map to estimate the position in the geodetic frame.
2. While the VIO suffers from drifts and the georegistration results can be inconsecutive and unsmooth, we propose to fuse the VIO and the georegistration results to achieve smooth, drift-free and precise geolocalization performance.
3. We conduct real flight experiments with different dynamics and in different visual environments. We also make comparisons with conventional methods to highlight the effectiveness of the proposed method.

The paper proceeds as follows: Section II reviews the related research. Section III presents the detailed approach. Sections IV and V demonstrate the experimental procedure and experimental results respectively, with the conclusions drawn in Section VI.

## II. RELATED WORK

The navigation system is the basic component that supports the UAV to complete complex tasks. In recent years, vision-based navigation techniques have received increasing focus. Existing visual navigation techniques can be broadly categorized as RVL and AVL [1].

VO [2] and VSLAM [13] are two widely used RVL approaches, where VO usually acts as a component in VSLAM. Vision-only relative localization algorithms include SVO [14], DSO [15] and ORB-SLAM [16]. To improve the accuracy and robustness of the vision system, other sensors have been proposed to integrate with the cameras. One of the most notable methods is to integrate IMUs with cameras, as IMU can provide a high rate, robust motion estimation and vision can provide relatively precise motion estimation. VINS [4] and ORB-SLAM3 [5] are representatives of the current visual-inertial techniques. VO and VSLAM can estimate the position of the vehicle and simultaneously reconstruct the position of the tracked features. However, RVL is not capable to estimate the absolute position relative to the geodetic frame unless the initial geodetic position and orientation are given. Moreover, VO suffers from accumulated errors and VSLAM requires loop closures to achieve accurate localization and mapping; this limits the application of UAV in situations where precise and efficient absolute navigation is required.

The AVL techniques aim to register the onboard camera images to a reference map to estimate the position. When the database is aligned with the geodetic frame, the geodetic position can then be obtained. For UAV applications, georeferenced maps captured by satellites [7][6][8][9][12] or other vehicles [17] have been widely used. Template matching is a long-developed technique to perform aerial imagery registration. Template matching methods treat the query camera image as a patch extract from the georeferenced map, therefore it firstly warps the image to a similar scale and orientation with the map, and then compares with different parts of the map. Normalized cross-correlation (NCC) [9][8], mutual information (MI) [18], normalized information distance (NID) [7] and other similarity parameters are used in template matching aerial image registration. Due to the computation costs of template similarity, an initial guess of the position is usually required to reduce the search space. As template matching methods are sensitive to orientation variations and horizontal angles, gimbals have been used to capture downward-looking camera images with fixed orientation [9][18][7].

An alternative AVL approach is to match the sparse features between the camera image and the reference map. Conte [19] used edge features for aerial image registration, but the results have shown low match rates. [10] proposed a novel feature named as abBRIEF, which was a modified version of BRIEF feature [20], for AVL implementation and has shown better performance than the conventional BRIEF feature. [17] used a self-constructed 3D landmark map as the reference, where the visual landmarks were described as BRISK features [20]; in the localization stage, they used weak GPS priors to reduce visual aliasing. [21] built a HOG feature lookup table, which contains HOG features extracted at every map pixel; their method requires no prior location information as the initial stage contains a global search process. Nassar [12] proposed to use SIFT features for aerial image registration, which is then followed with a semantic feature alignment to localize the vehicle. Bianchi [6] trained an autoencoder to describe the camera and map image; with the aid of prior location information, their method has shown a high localization rate with low localization error in different periods of a day. To reduce the visual aliasing, gimbals and known initials have also been used in feature-based aerial image registration methods [6][12][19].

In summary, VO suffers from drifts and VSLAM requires loop closures to achieve accurate localization and mapping performance, while at AVL it is hard to register every query camera image to the referenced map due to the visual appearance variations and the imperfect design of the match algorithms. In this paper, we propose to integrate visual-inertial odometry with a feature-based AVL component. The odometry can generate consecutive and smooth relative navigation results, while the AVL can transfer the relative navigation result to the geodetic frame and reduce the VIO drifts. Our method is closely related to [11], which not only integrated RVL and AVL in a factor graph, but also used 3D landmarks for registration. However, the authors in [11] used a tactical level IMU for dead reckoning and used a 3D model as the reference, which is expensive and hard to obtain. In this paper, we integrate a MIMU and a camera to achieve a low-cost and light-weighted navigation system. Furthermore, we only use a 2D map as the reference which is easier to obtain than the 3D model. Unlike most conventional methods that have used a gimbal and known initials, we work with unknown initials and rigidly mounted sensors; this can support the UAV applications in challenging enviroments and helps to integrate with the sensors distributed at different body parts of the vehicle.

## III. METHODOLOGY

The proposed method consists of three main components: the sliding-window VIO for relative navigation, the point registration for geolocalization and the pose graph for information fusion. The VIO integrates a MIMU and a monocular camera to estimate the motion states and reconstruct the 3D position of the tracked visual features in the local world frame. The point registration method firstly extracts map 2D features in an offline process; in the online localization process, it uses the reconstructed 3D points as the query point set, and registers the query points to the map points. The registered 3D points are then transferred to the geodetic frame and are used to relocate the vehicle with the PNP (Perspective-N-Point) algorithm. The pose graph optimization component fuses the relative navigation results from the VIO with the geodetic positions from the registration component to achieve consecutive and drift-free geolocalization performance.

### A. Relative Visual Localization with a Sliding Window VIO

We adapt the VINS-Mono [4], which is a state of the art visual-inertial odometry, as the RVL component in the proposed system. The VIO can estimate the UAV motion states (i.e. position, attitude and velocity) and simultaneously reconstructs the 3D position of the tracked points in the local world frame $W$. The full state vector of the VIO is defined as:

$$\chi = [x_0, x_1, ..., x_n, \lambda_0, \lambda_1, ..., \lambda_i]$$
$$x_k = [p_{B_k}^W, v_{B_k}^W, q_{B_k}^W, b_a, b_g], k \in [0, n] \quad (1)$$

where $x_k$ is the IMU body state at the k-th keyframe in the sliding window; it contains the position vector $p_{B_k}^W$, the velocity vector $v_{B_k}^W$, the quaternion $q_{B_k}^W$, the accelerometer bias $b_a$ and the gyroscope bias $b_g$. $\lambda_i$ is the inverse depth of the i-th feature tracked in the sliding window. Based on $\lambda_i$, the 3D position of the feature in the local world frame can be obtained as:

$$P_i^W = R_{B_k}^W \left( R_C^B \frac{1}{\lambda_i} \pi_c^{-1} \left( \begin{bmatrix} u_i^{C_k} \\ v_i^{C_k} \end{bmatrix} \right) + p_C^B \right) + p_{B_k}^W \quad (2)$$

where $[u_i^{C_k}; v_i^{C_k}]$ is the first observation of the i-th feature in the window; $\pi_c^{-1}$ is the back projection function that turns the pixel location into a unit vector using the camera intrinsic parameters. $R_C^B$ and $p_C^B$ are the extrinsic parameters between the IMU and the camera. In our implementation, we calibrated the extrinsic parameters offline and froze them during the navigation process. While we adopted VINS-Mono as the basic component, other keyframe-based VIO techniques [5][22] that can publish the required information, i.e. keyframe motion states, keyframe features and 3D reconstructed points, can also be used as the RVL component in the proposed system.

Note that, without the knowledge of the starting position, the VIO can only estimate the motion states related to the local world frame. Moreover, the integration process causes the VIO accuracy to degrade with the travel distance. In a VSLAM system, a loop closure component is used to detect previously explored areas, which is then used to relocate the vehicle and adjust the constructed map. However, loop closure only reduces the relative navigation errors and requires travelling back to previously visited places.

### B. Geolocalization by Registration of the VIO reconstructed Points to the Map

Prebuilt georeferenced maps contain rich features that can be used as landmarks for absolute visual navigation. Compared with the image registration research, geolocalization not only has to register the camera image features to the georeferenced map, but also has to estimate the position of the robot from the registration. However, visual appearance variations between the camera image and the map can fail the registration component; furthermore, the motion of the roll and pitch angles can cause the view field of the camera located distant to the downward area, which makes it difficult to estimate the UAV's precise position. To address the above challenge, we develop a novel point registration method for localization, which makes use of the VIO reconstructed 3D points, other than the conventional image 2D points, for geolocalization.

#### 1) Map Preprocessing

Offline map processing is the first step of the proposed aerial image registration component. As the onboard camera images and the reference maps can have very different qualities, the repetitiveness of the feature detector cannot be guaranteed. Therefore, we extract interest points with a constant stride (empirically set as 10 pixels in our experiments) in the map to ensure that the observed camera features are included in the map feature database; SIFT keypoints [23] are then constructed at the interest points with fixed orientation and size. Later in the online registration process, camera images are rotated and resized to have similar orientation and scale as the map with the aid of the orientation from the magnetic compass and the measured height from the VIO; then SIFT points are also built at the VIO tracked corner points with the same feature orientation and size as the map features to preserve the consistency. The offline map processing component generates a reference feature database represented as:

$$D = \{(f_j^m, l_j^m)\}_J \quad (3)$$

where $f_j^m$ is the feature descriptor to the j-th map feature, and $l_j^m$ is the 2D geodetic position of the j-th feature.

When anchoring the origin of the geodetic frame at the point corresponding to the map's left-up corner and define the geodetic frame as ENU (East-North-Up) orientation, $l_j^m$ can be expressed as:

$$l_j^m = \left[ u_j^m \cdot \sigma^m; -v_j^m \cdot \sigma^m \right] \quad (4)$$

where $\left[ u_j^m; v_j^m \right]$ is the detected feature position on the map and $\sigma^m$ is the resolution ratio of the map.

*2) Image Feature Extraction*

Compared to some existing works that use a separated feature registration process to the dead reckoning component [9][11][19], we propose to use the VIO tracked features for registration. This not only reduces the time consumption for image feature detection, but also helps to incorporate the 3D reconstruction information to augment the registration performance.

Instead of using all the feature points, we only select the points that have a reprojection error smaller than a predefined threshold for registration. If the number of the selected points is below a threshold, we skip the image frame. Once enough good reconstructed 3D points are available, the corresponding camera image is rotated with the aid of the magnetic compass, and rescaled to the map's scale based on the measured height from the VIO. The position of the tracked points on the new image can be calculated as:

$$\begin{bmatrix} u_i^{\bar{C}} \\ v_i^{\bar{C}} \end{bmatrix} = s R_{2D}\left(\theta_{mag}\right) \begin{bmatrix} u_i^C \\ v_i^C \end{bmatrix}$$
$$s = \frac{\sigma^m}{\sigma^{\bar{C}}} \approx \sigma^m \frac{f}{H} \quad (5)$$

where $\left[ u_i^C; v_i^C \right]$ is the feature observation at the camera image and $\left[ u_i^{\bar{C}}; v_i^{\bar{C}} \right]$ is the corresponding observation in the rotated and scaled image. $R_{2D}\left(\theta_{mag}\right)$ is the 2D rotation operation, which is related to the compass orientation $\theta_{mag}$. $s$ is the image scale factor, which is related to the resolution ratio of the map $\sigma^m$, the camera focal length $f$ and the VIO measured altitude $H$.

SIFT descriptors are then extracted from the rotated and scaled image with fixed orientation and scale as the same as the map features. The query point set corresponding to the k-th keyframe is represented as:

$$Q_k = \left\{ \left( f_i^C, P_i^W \right) \right\}_I \quad (6)$$

where $f_i^C$ is the SIFT descriptor and $P_i^W$ is the coordinate of the reconstructed 3D points as shown in (2).

Note that, while feature-based image registration methods and existing georeference map aided UAV localization approaches set the 2D points in the camera image as the query point set [6][7][9][10], the proposed method treats the reconstructed 3D points as the query set, as shown in (6).

*3) Point Registration and Geolocalization*

The proposed point registration process includes two steps: finding raw feature correspondence and filtering outliers. The raw matches are built by finding the correspondences based on the appearance similarity and the outlier filtering is achieved by fitting the matches to a model. Existing research that performs image-to-map registration [9][12][21][24] have adopted a model consisting of scale, rotation and translation parameters, where the translation parameter has been used to compensate the position drifts; however, this assumption is valid for cameras facing downward to the ground. For cameras with a rigid installation, the nonzero roll and pitch angles can lead to accuracy degradation. In this work, we focus on working with rigidly mounted sensors, as such installation helps to integrate with sensors distributed at other parts of the UAV.

Instead of aligning the camera image to the reference map, we propose to align the reconstructed 3D points to the map points in the geodetic horizontal level. Thanks to the use of IMU, the VIO can align with the gravity direction. By using a magnetic compass, we can correct the heading of the VIO and rotate the reconstructed points as:

$$P_i^G = R_W^G \left(\theta_{mag}, 0, 0\right) P_i^W \quad (7)$$

where $\theta_{mag}$ is the orientation measured by the compass corresponding to the first keyframe. $R(\cdot)$ translates the Euler angles to the rotation matrix.

After the rotation as shown in (7), the reconstructed points have similar direction and scale to the real world points, but their position drifts with the VIO errors. To compensate the position error, we propose to register the 3D points with the map at the horizontal level, which is modelled as a position translation vector. In detail, each matched pair can compute a 2D translation vector:

$$t_{i,j} = l_j^m - P_i^G(x, y) \quad (8)$$

where the j-th map feature and the i-th reconstructed point feature is a matched pair, and $P_i^G(x, y)$ means the 2D vector consists of the first two elements.

Then for every match, we count the number of the inlier correspondences, where a match is considered as an inlier if its translation vector has a variation smaller than a set threshold. For the match that has the most inliers and the inlier number is higher than an acceptance threshold, we consider it is a successful registration and average all the inliers' translation vectors for use. The translation vector is regarded as the position drift of the VIO reconstructed points. Therefore, for all the inliers, the obtained position translation vector is used to move the points to align with the geodetic frame as shown in Fig.1, and expressed as:

$$\bar{P}_i^G = P_i^G + t \quad (9)$$

where the first two elements of $t$ are the averaged 2D translation from all inliers, and the third element is zero. At this point, the inlier 3D points are moved to align with the geodetic frame in the horizontal plane. Then, the absolute position and attitude of the camera can be calculated by using the PNP algorithm between the camera points and the aligned 3D points. Then the absolute position of the body frame is expressed as:

$$p_{B_k}^G = p_{C_k}^G - R_{C_k}^G R_B^C p_C^B \quad (10)$$

where $p_{C_k}^G$ and $R_{C_k}^G$ are obtained from the PNP algorithm; $R_B^C$ and $p_C^B$ are the extrinsic calibration parameters between the IMU and the camera.

## C. Global Pose Graph Optimization

The mentioned VIO in section III.A can provide consecutive motion parameters with respect to the local world frame, while the georegistration in section III.B can provide geodetic position at the matched places. To achieve smooth and drift-free geolocalization performance, we propose to integrate the VIO and the point registration results to complement each other.

In the pose graph, each node represents the state vector containing the position and attitude of the vehicle, which is represented as:

$$n_k = \left[ p_{B_k}^G, q_{B_k}^G \right], k \in [0, n] \qquad (11)$$

The VIO outputs are used to construct the relative edges that describe the relative transformation between two nodes as:

$$\hat{p}_{ij}^{B_i} = \left( \hat{R}_{B_i}^W \right)^{-1} \left( \hat{p}_{B_j}^W - \hat{p}_{B_i}^W \right) \qquad (12)$$

$$\hat{q}_{ij} = \left( \hat{q}_{B_i}^W \right)^{-1} \otimes \hat{q}_{B_j}^W$$

where the hat represents the estimation from the VIO and $R$ represents the rotation matrix corresponding to the quaternion.

The residual factor can then be given as:

$$r_{ij}\left( q_{B_i}^G, p_{B_i}^G, q_{B_j}^G, p_{B_j}^G \right) = \begin{bmatrix} \left( R_{B_i}^G \right)^{-1} \left( p_{B_j}^G - p_{B_i}^G \right) - \hat{p}_{ij}^{B_i} \\ \left( q_{ij} \otimes \hat{q}_{ij}^{-1} \right)_{xyz} \end{bmatrix} \qquad (13)$$

where $(\cdot)_{xyz}$ extracts the vector part of a quaternion.

The geodetic positions from the aerial image registration are used to construct the absolute edges and the residual factor is expressed as:

$$r_i\left( p_{B_i}^G \right) = p_{B_i}^G - \hat{p}_{B_i}^G \qquad (14)$$

where $\hat{p}_{B_i}^G$ represents the position from the georegistration as shown in (10). Note that, the absolute edges are only available for the successfully registered frames.

The whole graph is optimized by minimizing the following cost function:

$$\min_{p^G, q^G} \left\{ \sum_{(i,j) \in S} \left\| r_{ij} \right\|_\Sigma^2 + \sum_{i \in L} \rho\left( \left\| r_i \right\|_\Omega^2 \right) \right\} \qquad (15)$$

where $S$ is the set of all the relative edge factors and $L$ is the set of all the absolute edges; $\Sigma$ and $\Omega$ is the information matrix of the edges, and $\rho(\cdot)$ is the Huber norm. The use of the Huber norm in (15) helps to reduce the impact of possibly wrong georegistrations.

## IV. EXPERIMENTAL SETUP

### A. System setup

The experiments were conducted by rigidly mounting the navigation system to a DJI M600Pro UAV as shown in Fig.2. The onboard sensors include a downward-looking monocular camera, a MIMU, a magnetic compass. A UBLOX-M8T GNSS receiver, which is reported to have a CEP error of 2 meters, was also mounted on the UAV to provide the position reference. The vehicle was also equipped with a polarized light compass for further research. Note that the polarized light compass was equipped at the top of the vehicle as shown in Fig.2.

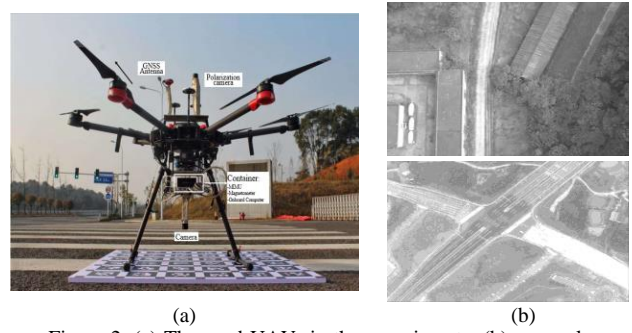

(a)          (b)
Figure 2. (a) The used UAV in the experiments; (b) a sample camera image from the Rural dataset displayed at the first row and a sample from the Testing Zone dataset displayed at the second row.

### B. Datasets

Two datasets were used to test the algorithms: The Rural dataset and the Testing Zone dataset. For each dataset, we conducted one flight for validation. The used 2D georeferenced maps were constructed by camera images captured by another UAV with a time separation longer than 3 months to simulate the visual variations. The used maps have a resolution ratio of 0.3 meter/pixel as shown in Fig.5. The geodetic frame was anchored at the left-up corner of the map, with the orientation defined as ENU as described in Section III. In the online navigation process, the camera worked at a frame rate of 10Hz, the MIMU worked at 200 Hz. The GNSS receiver worked at 10 Hz.

The Rural dataset covers an area of 0.51 km × 0.4 km located in Changsha, China. Most of the area along the flight trajectory is covered by natural features, such as trees and grassland, with a few buildings and roads as shown in Fig.5(a). The average height of this flight was about 110 meters above the ground. The flight lasted for about 6 minutes with a length of 1.18 kilometers. A sample camera image is shown in Fig.2(b). The resolution of the onboard camera is $640 \times 480$, with the field of view at about $41° \times 32°$.

The Test Zone dataset contains an area of 1.46 km × 1.45 km at the National Intelligent Connected Vehicle (Changsha) Testing Zone. The Zone contains city subzones, highway subzones and urban subzones as shown in Fig.5(b). The average height of this flight was about 285 meters, and the flight lasted for about 7 minutes with a length of 2.45 kilometers. A sample camera image is shown in Fig.2(b). The resolution of the onboard camera is $640 \times 512$, with the field of view at about $67° \times 56°$.

The details of the two datasets are shown in Table.I. Note that, the two flights captured images of various natural visual appearances with different fly speeds and heights, which make them valid to test the algorithm in different conditions.

TABLE. I  DETAIL OF THE CONDUCTED FLIGHTS

| Dataset | Time (s) | Length (km) | Avg Hight (m) | Avg Horizontal Angle (°) |
|---|---|---|---|---|
| *Rural* | 448.20 | 1.18 | 110 | 5.63 |
| *Testing Zone* | 407.50 | 2.45 | 285 | 7.80 |

### C. Experimental Procedure

For algorithm validation, we empirically extracted the map points with a constant 10-pixels' stride and detect the camera image points as described in [4]. In all, there were 22,042 map features in the Rural dataset and 116,354 map features in the Testing Zone dataset. To speed up the feature matching process, we used the FLANN-based matcher [25] to find the best match for every query feature.

Two georegistration methods were used for comparison. The first method (M1) performs image-level registration, where an image translation was searched to best align the rotated and scaled camera image with the map; this obtained image translation was then mapped to position translation to compensate for the localization drifts. Such a registration strategy for geolocalization has been used in both the template matching methods [8][18] and the feature matching methods [12][24]. The second method (M2) also uses the same registration process, but SIFT feature detector was used at the front end to find interest points and automatically set the SIFT feature parameters for both the camera images and the map image; this method is consistent with conventional image registration method where feature detectors have been used to automatically find interest points in both the query and the reference images [12][24]. The details of the proposed method and the two conventional methods are shown in Table II.

Table III shows some key parameter settings of the algorithm. The listed parameters were set empirically but were set the same for the two datasets to keep consistency.

TABLE. II DETAIL OF THE CONDUCTED GEOREGISTRATION METHODS

| Method | Feature Detector | Feature Descriptor | Registration Method |
|---|---|---|---|
| M1 | Camera: Shi-Tomasi corner feature Map: extract with every 10 pixels' stride | SIFT descriptor with fixed orientation and size | Align the camera image with the map image |
| M2 | SIFT detector for both the camera images and the map images | SIFT descriptor with parameters set by the detector | Align the camera image with the map image |
| Proposed | Camera: Shi-Tomasi corner feature Map: extract with every 10 pixels' stride | SIFT descriptor with fixed orientation and size | Align the reconstructed 3D points to the map points in the geodetic horizontal level |

TABLE. III THE DETAIL OF THE USED PARAMETERS

| Parameter | Value |
|---|---|
| map point stride | 10 map pixels (3 meters) |
| 3D point selection threshold | reprojection error ≤ 8.0 |
| valid potential match keyframe | Observe more than 20 selected points |
| translation inlier threshold | 30 map pixels (9 meters) |
| success match threshold | 15 inlier matches |

## V. EXPERIMENTAL RESULTS

Note that the VIO only estimates the relative pose, while the georegistration component and the fusion component can estimate the absolute geolocation. For a fair comparison, we aligned the VIO results with the ground truth by using the *Umeyama* algorithm [26] as the widely used ATE [27] evaluation method does and then used the aligned VIO poses for evaluation. As the VIO component has been studied comprehensively in [4], here we focus on analysing the performance of the georegistration component and the fusion component.

### A. Aerial Image Registration Performance

The aerial image registration performance of the three conducted methods is presented in this section. Fig.3 shows the true match rate of the three methods, where a true match requires at least eight inlier matches and the registered place is within 30 meters to the ground truth location.

In both the Rural dataset and the Testing Zone dataset, M1 shows a higher match rate than M2. The premise of a successful image registration requires the detector to find out the same features in both the query and the reference images with similar feature parameters. However, the camera and map images were captured at different times with different cameras; this is challenging for the SIFT feature detector to detect consistent features in the two source images. Therefore, M2 shows a lower true match rate than M1 in both the Rural and the Testing Zone datasets due to the usage of the automatic feature detector, as described in Table.II.

Though the proposed method and M1 use the same features for image registration, they show different performances as shown in Fig.3. In the Rural dataset, the vehicle flew at a low altitude and the roll and pitch angles are close to zero as shown in Table.II; in this situation, the relation between the scaled and rotated camera images and the map image can be regarded as a pure image translation, and M1 can achieve a high match rate. Moreover, as M1 uses all the camera image features, it achieves a slightly better performance than the proposed method, which only selects the points with a small reprojection error for use. In the Testing Zone dataset, the flight height, speed, and horizontal angles are larger than those in the Rural dataset; this is more challenging for M1 as pure translation is not adequate to describe the relations between the camera images and the map image. Benefit from using the VIO reconstructed 3D points, which is horizontally aligned with the geodetic frame, the proposed method achieves a much higher match rate than M1.

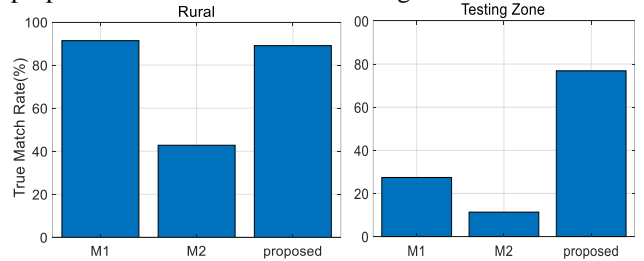

Figure 3. The match rate of the three conducted methods in the Rural and the Testing Zone datasets. The proposed method shows a consistent high match rate in both datasets.

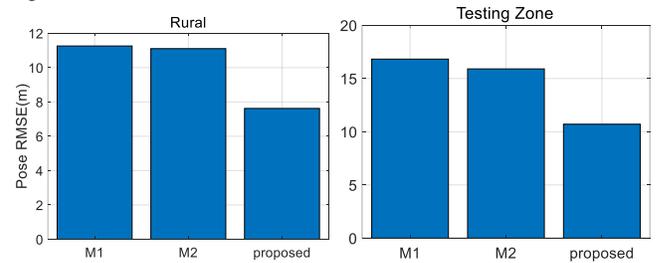

Figure 4. The geolocalization error in the Rural and the Testing Zone datasets. The proposed method achieves the best performance in both datasets.

Fig.4 shows the position root mean square error (RMSE) of the true matches of the three conducted methods. As both M1 and M2 perform image-level registration, their accuracy is comparable in the two datasets. We highlight that the proposed aerial image registration method shows the best localization precision in both the Rural and the Testing Zone datasets, which demonstrates the performance improvements brought in by using the reconstrued 3D points. Note that, the proposed method's localization error in the Testing Zone dataset is larger than that in the Rural dataset; this is mainly since the flight height in the Testing Zone dataset is higher, which is more challenging for precise 3D reconstruction.

In summary, Fig.3 and Fig.4 show that the proposed registration method has a consistent high match rate and the smallest localization error.

We also investigated the time consumption of the image registration component in a computer with a 2.60 GHz Intel i7-6700HQ processor and 16 GB RAM. The code was run as a ROS node. In the Rural dataset, the average georegistration time was about 15.6 milliseconds per frame; in the Testing Zone dataset, it took about 216.3 milliseconds to process one frame. Further improvements, such as using the previous localization result to bound the feature search area, can reduce the time consumption.

### B. Localization Performance by Fusing the VIO with the Georegistration

The final localization output of the proposed system is obtained by fusing the relative localization result from the VIO and the geodetic position from the registration component. To highlight the advantage of fusing the relative and absolute localization techniques, we present the precision of the proposed fusion method and the precision of its components in Table.IV. As the initial position is unknown, the vehicle is not aware of its geodetic location before the first success georegistration; but once a georegistration is detected, the vehicle's trajectory is transferred to align with the geodetic frame.

TABLE.IV LOCALIZATION PERFORMANCE

| Dataset | VIO Pose RMSE(m) | Registration Pose RMSE(m) | Fusion Pose RMSE(m) |
|---|---|---|---|
| Rural | 18.58 | 7.41 | **3.76** |
| Testing Zone | 69.70 | 10.70 | **8.75** |

It can be seen from Table.IV that the VIO has the largest localization error in the two datasets, as its errors accumulate in the integration process. The absolute position error obtained from the georegistration is much smaller than the VIO and it is drift-free. Fig.5 shows the registered places in the Rural and the Testing Zone datasets. The registered places distribute along the ground truth trajectory; however, the distribution is not smooth and there are also some wrong registrations distant from the ground truth trajectory. Such unsmooth and false localization results can bring in risks for the UAV control.

By fusing the relative and absolute localization results as described in section III.C, we achieve the best localization results. In the Rural dataset, the fused position error is 3.76 meters; while in the Testing dataset, the fused position error is 8.75 meters. The localization error from the pose graph optimization is much smaller than the VIO, as the pose graph incorporates with the georegistration to fix the VIO drifts. Moreover, the fused results are also more accurate than the georegistration results; this is a benefit from adding the relative constraints into the pose graph to achieve smooth localization performance. Note that, the wrong registration results have little influence on the optimized results, which demonstrates the advantage of using Huber loss in the objective functions as shown in (15).

In summary, by fusing the relative localization results from the VIO and the absolute localization results from the georegistration component, we obtain smooth, drift-free and precise geolocalization results. The proposed method provides a novel solution to GNSS-denied autonomous localization.

### C. Compared with Other Works

This part compares the experimental results with some other research published in recent years. Due to the lack of publicly available code and datasets, we only summarize the reported localization performance and the corresponding experimental conditions. The details of the comparison are shown in Table.V.

It can be seen from Table.V that [9][11] and our method have integrated a dead reckoning component with the georegistration component to achieve consecutive and drift-free localization performance, while other work only focuses on georegistration. Note that, [9] used the Yamaha Rmax UAV, which is embedded with a relatively good IMU, and [11] used tactical-level IMU for dead reckoning, while the proposed method combines a light-weighted MIMU with a camera to form a VIO. Note that, both our method and [11] incorporate 3D information in geolocalization. However, [11] requires a 3D map and known initial positions, while the proposed method only uses a 2D map with unknown initials. Although it works with MIMU and 2D maps, the proposed method shows higher localization precision than the reported results in [9] and [11].

Our method's localization accuracy in the Rural dataset is slightly lower than the results shown in [6] and [7]. However, we mounted the sensors rigidly to the UAV body, which is more challenging than using a gimbal based camera installation. Without unknown initials, the proposed method achieves comparable localization accuracy with [6] and [7], where a known initial position is required.

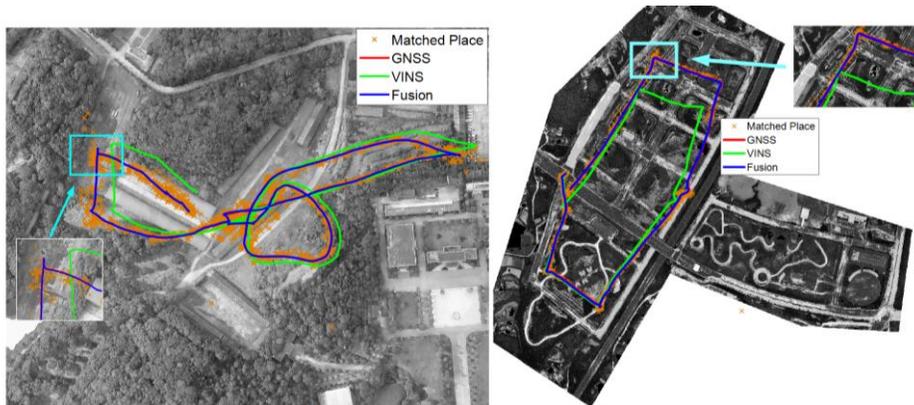

Figure 5. Demonstration of the localization results of the Rural (left) and the Testing Zone (right) datasets by plotting on the map. By fusing the relative localization results from the VIO and the absolute localization results from the aerial image registration, the proposed method achieves consecutive, drift-free and precise geolocalization performance.

TABLE.V COMPARISION WITH STATE OF THE ART WORK

| Author & Year | Method | Experiments | Known initial | Accuracy |
|---|---|---|---|---|
| Conte et al. 2009 [9] | image registration + Kalman filter | Flight length: 1km; Flight height: 60 m; Sensors: good IMU and camera in a **gimbal**; Map: 2D map; | Yes | <8m |
| Chiu et al. 2014 [11] | 3D point registration + factor graph | Flight length: 26.5km; Flight height: --; Sensors: good IMU and camera; Map: **3D map** | Yes | 9.35 m |
| Shan et al. 2015 [21] | image registration | Flight length: 0.72 km; Flight height: --;Sensor: camera; Map: 2D map | **No** | 6.77 m |
| Nasser et al. 2018 [12] | image registration | Flight length: --; Flight height: --; Sensor: camera; Map: 2D map | Yes | 10.4 m |
| Mantelli et al. 2019[10] | image registration | Flight length: 2.4km; Flight height: --; Sensor: camera in a **gimbal**; Map: 2D map | Yes | 17.78 m |
| Patel et al. 2020 [7] | image registration | Flight length: 1.1km; Flight height: 36-42m; Sensor: camera in a **gimbal;** Map: 2D map | Yes | $< 3m$ |
| Bianchi et al. 2021 [6] | Image registration | Flight length: 1.1km; Flight height: 36-42m; Sensor: camera in a **gimbal**; Map: 2D map | Yes | $< 3m$ |
| **Ours** | 3D points alignment + pose graph | Flight length: 1.18/2.45km; Flight height: **~110/285m;** Sensor: **MIMU** and camera with **rigid installation;** Map: 2D map | **No** | **3.76/ 8.75 m** |

## VI. CONCLUSIONS AND DISCUSSIONS

In this paper, we have proposed a novel UAV localization method by using a camera, a MIMU, a magnetic compass, and a 2D georeferenced map. The method consists of three components: the visual-inertial odometry for relative localization, the point registration for geolocalization and the pose graph for information fusion. The VIO provides relative motion constraints to the pose graph, while also reconstructs the 3D position of the observed features. Other than performing image-level registration, the proposed method aligns the reconstructed 3D points to the map points in the geodetic frame, which has shown a consistent high registration rate and a low geolocalization error. After obtaining the relative position from VIO and the geodetic position from the registration component, the pose graph component fuses them to achieve consecutive, drift-free and precise localization performance. Compared to existing work, the proposed method requires no prior knowledge of the initial position and achieves precise localization without using a gimbal; such a rigid sensor installation helps to fuse with sensors distributed at other parts of the vehicle.

Future work can focus on using the odometry to bound the feature matching area, which can speed up the registration process and improve the registration success rate. We are also planning to update the onboard UAV system to include a GNSS RTK device to provide centimeter level ground truth.